\documentclass[runningheads]{llncs}
\usepackage[misc,geometry]{ifsym}
\usepackage{graphicx}
\usepackage{todonotes}
\usepackage{booktabs}
\usepackage{bigdelim}
\usepackage[strings]{underscore}

\usepackage{amsmath}
\usepackage{amssymb}
\usepackage{mathtools}
\usepackage{namedtensor}
\usepackage[symbol]{footmisc}

\usepackage{ifthen}
\newboolean{arxiv}
\setboolean{arxiv}{true}

\usepackage{lineno}
\usepackage{xcolor}
\definecolor{applica}{HTML}{133e80}
\definecolor{aorange}{HTML}{EA6B66}
\definecolor{agreen}{HTML}{67AB9F}
\definecolor{agray}{HTML}{808080}

\begin{document}
\title{Going Full-TILT Boogie on Document Understanding with Text-Image-Layout Transformer}
\titlerunning{Text-Image-Layout Transformer}
% DOcument InteliGEnce Transformer
\newcommand*\samethanks[1][\value{footnote}]{\footnotemark[#1]}
\newcommand*\samethanksa[1][\value{footnote}]{\footnotemark[#1]}

\ifthenelse{\boolean{true}}{
    \author{Rafał Powalski\inst{1, \thanks{RP, ŁB contributed equally.} } \and 
    Łukasz Borchmann\inst{1,2, \samethanks} (\Letter) \and
    Dawid Jurkiewicz\inst{1,3, \thanks{DJ, TD, MP contributed equally.}} \and
    Tomasz~Dwojak\inst{1,3,\samethanksa}\and
    Michał Pietruszka\inst{1,4,\samethanksa}\and
    Gabriela Pałka\inst{1,3}
    %\and Second Author\inst{1} \and Third Author\inst{1}
    }
    \authorrunning{Powalski et al.}
    \institute{\textsuperscript{\rm 1} Applica.ai, Warsaw, Poland\\
    \email{name.surname@applica.ai}\\
    \textsuperscript{\rm 2} Poznan University of Technology, Poznań, Poland\\
    \textsuperscript{\rm 3} Adam Mickiewicz University in Poznań, Poland\\
    \textsuperscript{\rm 4} Jagiellonian University, Cracow, Poland    
    }
}{
    \author{Anonymous.}
}
\maketitle              % typeset the header of the contribution
\begin{abstract}
%MP draft:
%
%While nowadays NLP concentrates its efforts on understanding text, richly-formatted documents remain a nemesis, with their pletorous forms and erroneous OCR representations. Our solution to this problem unifies advances in Layout-Aware, Encoder-Decoder, and Multimodal approaches into an end-to-end trainable model that achieves state-of-the-art results in Visual Question Answering, Key Information Extraction and Document Classification.

We address the challenging problem of Natural Language Comprehension beyond plain-text documents by introducing the TILT neural network architecture which simultaneously learns layout information, visual features, and textual semantics. Contrary to previous approaches, we rely on a decoder capable of unifying a variety of problems involving natural language. The layout is represented as an attention bias and complemented with contextualized visual information, while the core of our model is a pretrained encoder-decoder Transformer.
 Our novel approach achieves state-of-the-art results in extracting information from documents and answering questions which demand layout understanding (DocVQA, CORD, SROIE). At the same time, we simplify the process by employing an end-to-end model.
%Top-class performance was proven by winning ICDAR Robust Reading Challenge for 2021.

\end{abstract}
\section{Introduction}

Most tasks in Natural Language Processing (NLP) can be unified under one framework by casting them as triplets of the question, context, and answer~\cite{pmlr-v48-kumar16,DBLP:journals/corr/abs-1806-08730,khashabi2020unifiedqa}. We consider such unification of Document Classification, Key Information Extraction, and Question Answering in a demanding scenario where context extends beyond the text layer. %In particular, when there is a need to include the spatial relationship between words and visual elements, such as plots, tables, and diagrams.
This challenge is prevalent in business cases since contracts, forms, applications, and invoices cover a wide selection of document types and complex spatial layouts.

% Nowadays, research in Natural Language Processing goes beyond plain-text problems and poses advanced challenges at the intersection with other fields such as Computer Vision (CV).
% Processing business documents such as contracts, agreements, applications, and invoices is an example of such a problem.
% In such scenario, information exceeds text layer and include layout (e.g. spatial relationship between words) and visual elements (e.g. plots, tables, diagrams).
% Low quality of OCR, wide selection of document types, and complex spatial layout make the task demanding, therefore a system cannot rely solely on text but requires incorporating information from structure and image.

\subsubsection{Importance of Spatio-Visual Relations.}  
The most remarkable successes achieved in NLP involved models that map raw textual input into raw textual output, which usually were provided in a digital form.
An important aspect of real-world oriented problems is the presence of scanned paper records and other analog materials that became digital.

Consequently, there is no easily accessible information regarding the document layout or reading order, and these are to be determined as part of the process. Furthermore, interpretation of shapes and charts beyond the layout may help answer the stated questions. A system cannot rely solely on text but requires incorporating information from the structure and image.

\begin{figure}[htp]
\centering
\resizebox{0.9\textwidth}{!}{ %%begin resize box
\includegraphics[width=.3\linewidth]{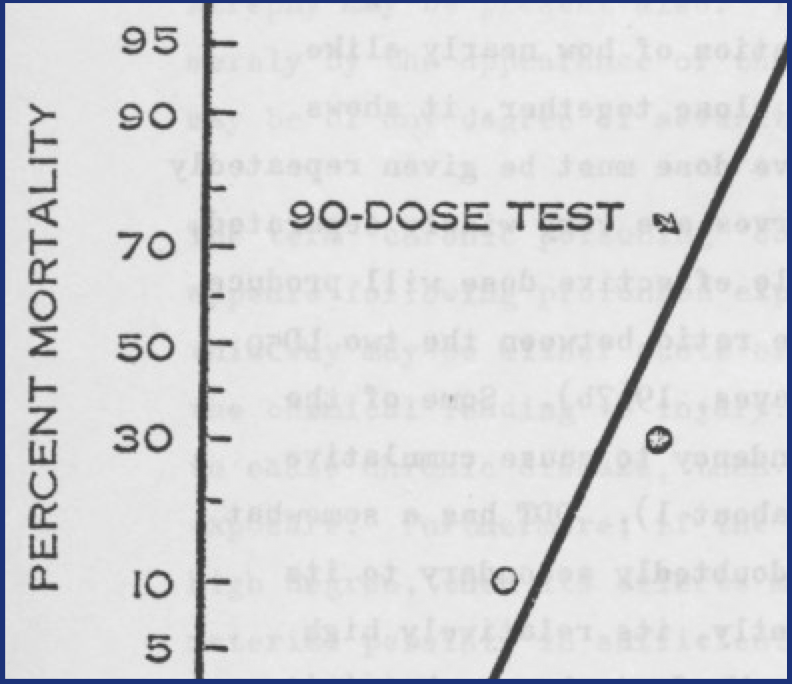}\hspace{2mm}
\frame{\includegraphics[width=.3\linewidth]{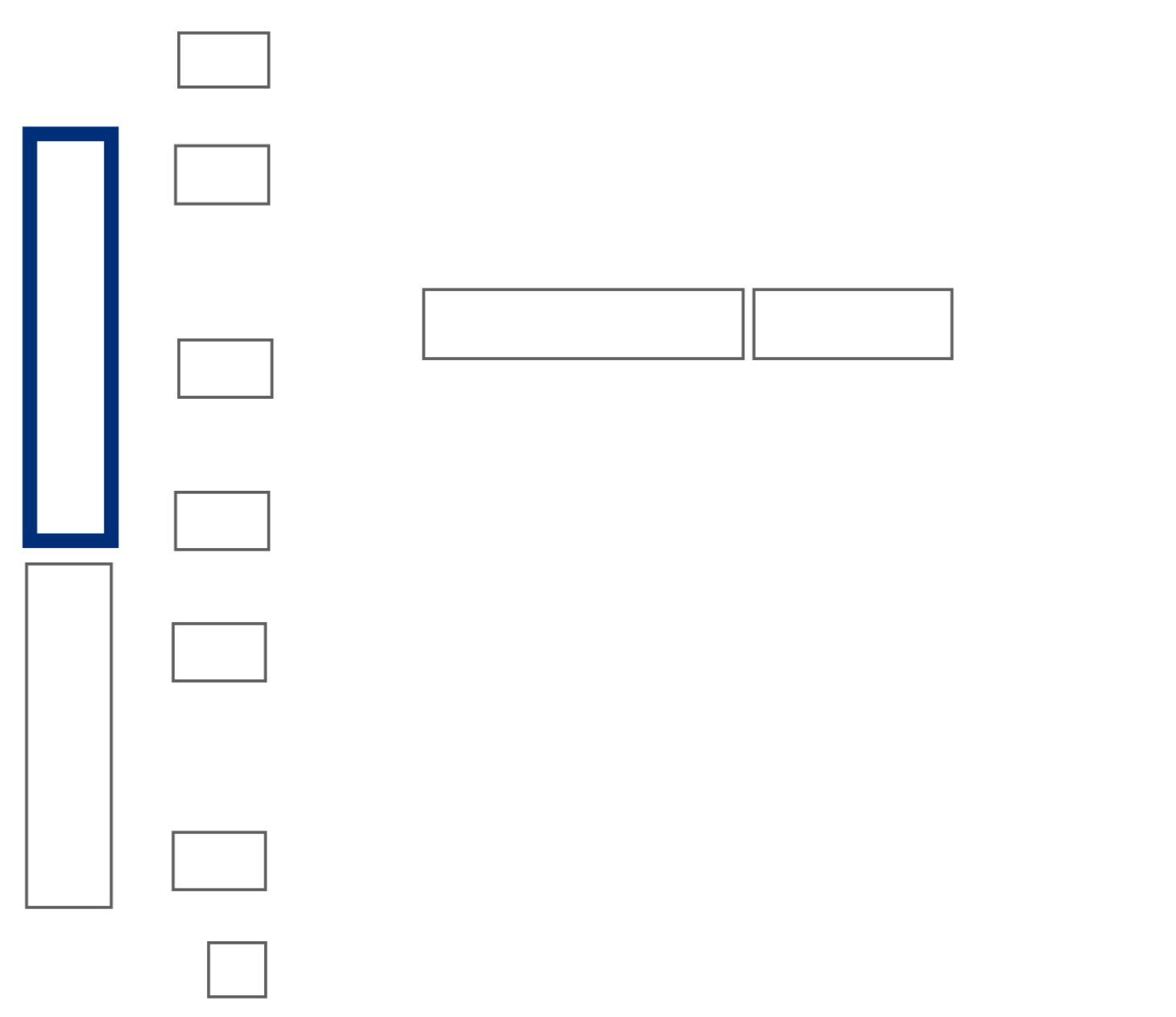}}\hspace{2mm}
\framebox(105,90){\parbox{80pt}{95 90 PERCENT\\{\color{applica}{\textbf{MORTALITY}}} 70\\ 90-DOSE TEST\\70 50 30 10 5}}
}%%
\caption{The same document perceived differently depending on modalities. Respectively: its visual aspect, spatial relationships between the bounding boxes of detected words, and unstructured text returned by OCR under the detected reading order.
%The form on the left focuses on the visual aspect. The schema in the middle presents spatial relationships between the bounding boxes of detected words. Finally, there is an unstructured text of the same excerpt on the right, as returned by OCR under the detected reading order.
}
\label{fig:modalities}
\end{figure}

Thus, it takes three to solve this fundamental challenge --- the extraction of key information from richly formatted documents lies precisely at the intersection of NLP, Computer Vision, and Layout Analysis (Figure~\ref{fig:modalities}).
These challenges impose extra conditions beyond NLP that we sidestep by formulating layout-aware models within an encoder-decoder framework.

%Intractable noise characterizing the conversion from physical to electronic format is also a phenomenon strictly observed on the applied side of the NLP research.

% MP - pracuję nad tym

\subsubsection{Limitations of Sequence Labeling.}
Sequence labeling models can be trained in all cases where the token-level annotation is available or can be easily obtained. Limitations of this approach are strikingly visible on tasks framed in either key information extraction or property extraction paradigms~\cite{Huang2019ICDAR2019CO,dwojak-etal-2020-dataset}. Here, no annotated spans are available, and only property-value pairs are assigned to the document. Occasionally, it is expected from the model to mark some particular subsequence of the document. However, problems where the expected value is not a substring of the considered text are unsolvable assuming sequence labeling methods\ifthenelse{\boolean{arxiv}}{ (Table~\ref{tab:comparison}).}{.\footnote{Expected values have always an exact match in CoNLL, but not elsewhere, e.g., it is the case for 20\% WikiReading, 27\% Kleister, and 93\% of SROIE values.}} As a result, authors applying state-of-the-art entity recognition models were forced to rely on human-made heuristics and time-consuming rule engineering.

\ifthenelse{\boolean{arxiv}}{Particular problems one has to solve when employing a sequence-labeling method can be divided into three groups. We investigate them below to precisely point out the limitations of this approach.}{}

\ifthenelse{\boolean{arxiv}}{
\begin{table}[t]
    \setlength{\tabcolsep}{10pt}
    \centering
    \begin{tabular}{llccc}
        \toprule
        Task &
        Annotation &
        Exact match &
        Layout \\
        \midrule
        CoNLL~2003 & word-level & 100\% & $-$ \\
        SROIE & \hspace{-11pt}\rdelim\}{3}{30pt}[ document-level] & 93\% & $+$ \\
        WikiReading & & 20\% & $-$ \\
        Kleister & & 27\% & $+$ \\
        \bottomrule  \\
    \end{tabular}
    \caption{Comparison of
    extraction tasks. Expected values are always present in a substring of a document in NER, but not elsewhere. Our estimation. %In the case of Kleister and SROIE percentage depends on used OCR and the reading-order it detected.
    \label{tab:comparison}}
\end{table}
}{}

Take, for example, the total amount assigned to a receipt in the SROIE dataset~\cite{Huang2019ICDAR2019CO}. Suppose there is no exact match for the expected value in the document, e.g., due to an OCR error, incorrect reading order or the use of a different decimal separator. Unfortunately, a sequence labeling model cannot be applied off-the-shelf. Authors dealing with property extraction rely on either manual annotation or the heuristic-based tagging procedure that impacts the overall end-to-end results~\cite{xu2020layoutlm,gralinski2020kleister,garncarek2020lambert,hong2021bros,xu2020layoutlmv2,liu2019graph}. Moreover, when receipts with one item listed are considered, the total amount is equal to a single item price, which is the source of yet another problem. Precisely, if there are multiple matches for the value in the document, it is ambiguous whether to tag all of them, part or none.
%\red{\paragraph{Tagging.} A sequence labeling model cannot be applied off-the-shelf due to OCR error, incorrect reading order, or normalization mismatch between document and answer. The precise cause is either the lack of an exact answer in the document or having a repeated answer. While either manual annotation or the heuristic-based tagging procedure may be employed, they negatively impact the overall end-to-end results.}

%\paragraph{From text spans to expected value.}
Another problem one has to solve is which and how many of the detected entities to return, and whether to normalize the output somehow. Consequently, the authors of Kleister proposed a set of handcrafted rules for the final selection of the entity values~\cite{gralinski2020kleister}.
These and similar rules are either labor-intensive or prone to errors~\cite{8270005}.

%\paragraph{Inferable properties.}
Finally, the property extraction paradigm does not assume the requested value appeared in the article in any form since it is sufficient for it to be inferable from the content, as in document classification or non-extractive question answering~\cite{dwojak-etal-2020-dataset}. %These are not solvable given sequence labeling architectures and require different treatment.%\\\noindent

\subsubsection{Resorting to Encoder-Decoder Models.}
Since sequence labeling-based extraction is disconnected from the final purpose the detected information is used for, a typical real-world scenario demands the setting of Key Information Extraction.

%As mentioned in prevSequence labeling models are disconnected from the final purpose the extracted information is used for.
To address this issue, we focus on the applicability of the encoder-decoder architecture since it can generate values not included in the input text explicitly~\cite{hewlett-etal-2016-wikireading} and performs reasonably well on all text-based problems involving natural language~\cite{2020t5}. Additionally, it eliminates the limitation prevalent in sequence labeling, where the model output is restricted by the detected word order, previously addressed by complex architectural changes (Section~\ref{sec:related_works}).

Furthermore, this approach potentially solves all identified problems of sequence labeling architectures and ties various tasks, such as Question Answering or Text Classification, into the same framework. For example, the model may deduce to answer \textit{yes} or \textit{no} depending on the question form only. Its end-to-end elegance and ease of use allows one to not rely on human-made heuristics and to get rid of time-consuming rule engineering required in the sequence labeling paradigm. 

Obviously, employing a decoder instead of a classification head comes with some known drawbacks related to the autoregressive nature of answer generation. This is currently investigated, e.g., in the Neural Machine Translation context, and can be alleviated by methods such as lowering the depth of the decoder~\cite{ren-etal-2020-study,kasai2020deep}. However, the datasets we consider have target sequences of low length; thus, the mentioned decoding overhead is mitigated.

\bigskip\noindent The specific contribution of this work %includes the unique extension of the Transformer to multi-modal input, use of encoder-decoder to achieve the robustness and independence from sequential order, a novel masking method, application of image embeddings already contextualized on multiple resolution levels, as well as 2D bias used instead of positional embeddings. These 
can be better understood in the context of related works.

\section{Related Works}\label{sec:related_works}

We aim to bridge several fields, with each of them having long-lasting research programs; thus, there is a~large and varied body of related works. We restrict ourselves to approaches rooted in the architecture of Transformer~\cite{transformer} and focus on the inclusion of spatial information or different modalities in text-processing systems, as well as on the applicability of encoder-decoder models to Information Extraction and Question Answering.

% CZASOWO UKRYWAM ;)
% \begin{figure}
%     \centering
%     \includegraphics[width=1\textwidth]{images/model/model_schema.pdf}
%     \caption{Our model incorporates document layout, text and raw pixels to produce the answer.}
%     \label{fig:model}
% \end{figure}.

\begin{figure}[t]
    \centering
    \includegraphics[width=0.45\textwidth]{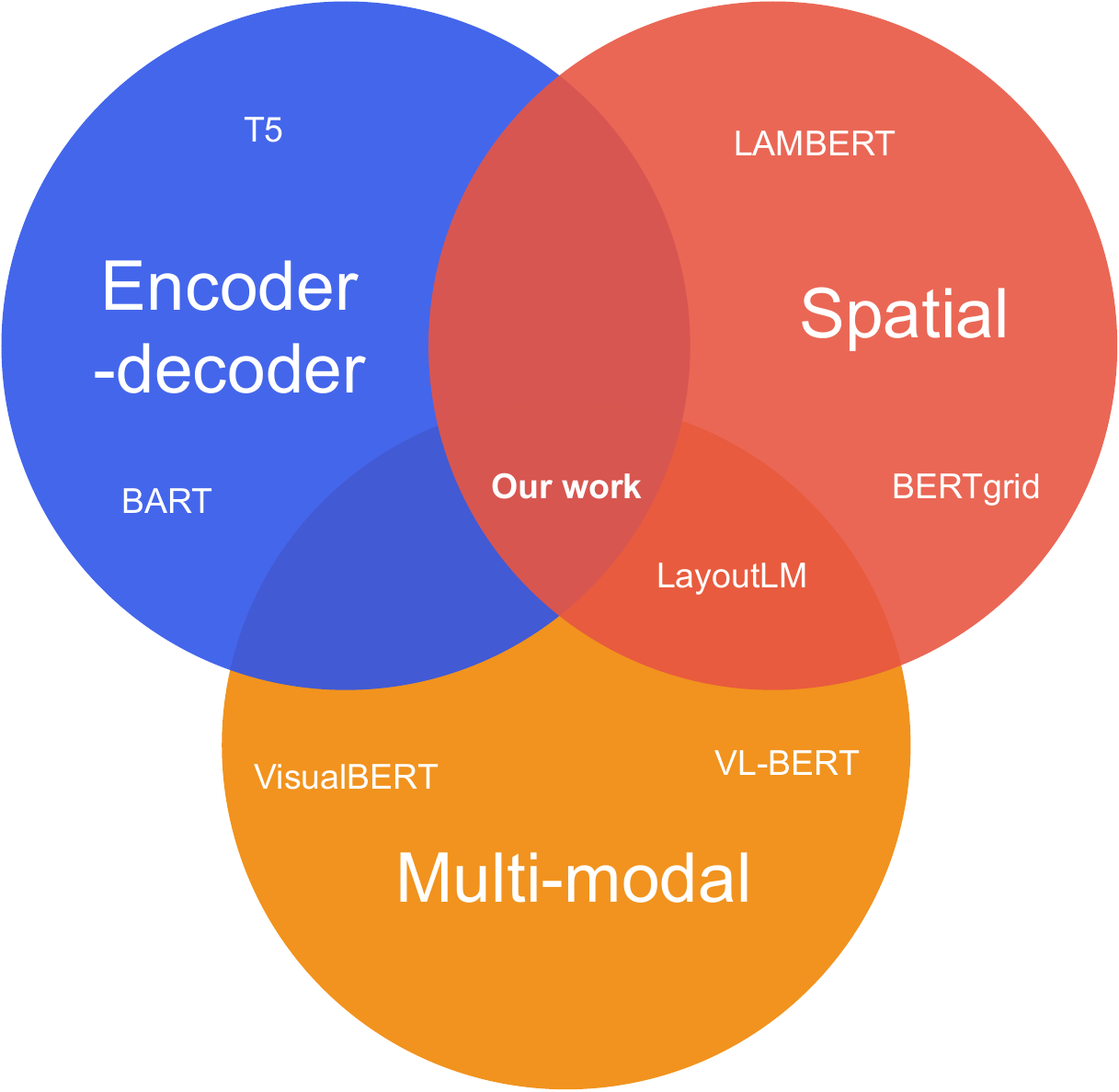}
    \caption{Our work in relation to encoder-decoder models, multi-modal transformers, and models for text that are able to comprehend spatial relationships between words.}
    \label{fig:venn}
\end{figure}

\subsubsection{Spatial-aware Transformers.} 
Several authors have shown that, when tasks involving 2D documents are considered, sequential models can be outperformed by considering layout information either directly as positional embeddings ~\cite{ho2019axial,garncarek2020lambert,xu2020layoutlm} or indirectly by allowing them to be contextualized on their spatial neighborhood ~\cite{denk2019bertgrid,yin2020tabert,herzig2020tapas}. 
Further improvements focused on the training and inference aspects by the inclusion of the area masking loss function or achieving independence from sequential order in decoding respectively~\cite{hong2021bros,hwang2020spatial}.
In contrast to the mentioned methods, we rely on a bias added to self-attention instead of positional embeddings and propose its generalization to distances on the 2D plane. Additionally, we introduce a novel word-centric masking method concerning both images and text. 
Moreover, by resorting to an encoder-decoder, the independence from sequential order in decoding is granted without dedicated architectural changes.
%Authors of the BERTgrid~\cite{denk2019bertgrid} have shown that, when tasks involving 2D documents are considered, previous spatial and sequential models can be outperformed by considering a grid of contextualized word piece embedding vectors with positions based on their neighborhood within the source document.Concurrent work introducing LayoutLM~\cite{xu2020layoutlm} and LAMBERT~\cite{garncarek2020lambert} models injected information about words' 2D coordinates to language models through additional positional embeddings, leading to spatial-aware self-attention mechanisms.A similar approach was introduced in BROS~\cite{hong2021bros} model along with area masking loss function. In~\cite{hwang2020spatial} authors proposed SPADE---a spatial-aware decoding algorithm for sequence tagging. Moreover, two-dimensional transformers were previously considered in the context of relationships between image regions. In particular, Axial Transformer offers a generalization of self-attention mechanism that, among others, utilize positional embeddings learned for each axis separately, which resemble LayoutLM's method of embedding spatial information~\cite{ho2019axial}. Another line of work also occupied with 2D structures addresses understanding of tabular data. To deal with spatial dependencies between table cells the authors of TaBERT~\cite{yin2020tabert} use vertical self-attention over rows, while proponents of TAPAS~\cite{herzig2020tapas} introduce additional summable embeddings that capture tabular structure.

\subsubsection{Encoder-decoder for IE and QA.} Most NLP tasks can be unified under one framework by casting them as Language Modeling, Sequence Labeling or Question Answering~\cite{radford2019language,Keskar2019UnifyingQA}. The QA program of unifying NLP frames all the problems as triplets of question, context and answer~\cite{pmlr-v48-kumar16,DBLP:journals/corr/abs-1806-08730,khashabi2020unifiedqa} or item, property name and answer~\cite{hewlett-etal-2016-wikireading}. Although this does not necessarily lead to the use of encoder-decoder models, several successful solutions relied on variants of Transformer architecture~\cite{transformer,lewis-etal-2020-bart,dwojak-etal-2020-dataset,2020t5}. The T5 is a prominent example of large-scale Transformers achieving state-of-the-art results on varied NLP benchmarks~\cite{2020t5}. 
We extend this approach beyond the text-to-text scenario by making it possible to consume a multimodal input.

\subsubsection{Multimodal Transformers.}
The relationships between text and other media have been previously studied in Visual Commonsense Reasoning, Video-Grounded Dialogue, Speech, and Visual Question Answering~\cite{han2021survey,le-etal-2019-multimodal,DBLP:journals/corr/abs-1910-11559}. In the context of images, this niche was previously approached with an image-to-text cross-attention mechanism, alternatively, by adding visual features to word embeddings or concatenating them~\cite{9037732,lee2018stacked,li2019visualbert,Su2020VL-BERT,xu2020layoutlm}. 
We differ from the mentioned approaches, as in our model, visual features added to word embeddings are already contextualized on an image's multiple resolution levels~(see Section \ref{sec:arch}). 

\ifthenelse{\boolean{arxiv}}{
    \begin{figure}
        \centering
        \includegraphics[width=0.9\linewidth]{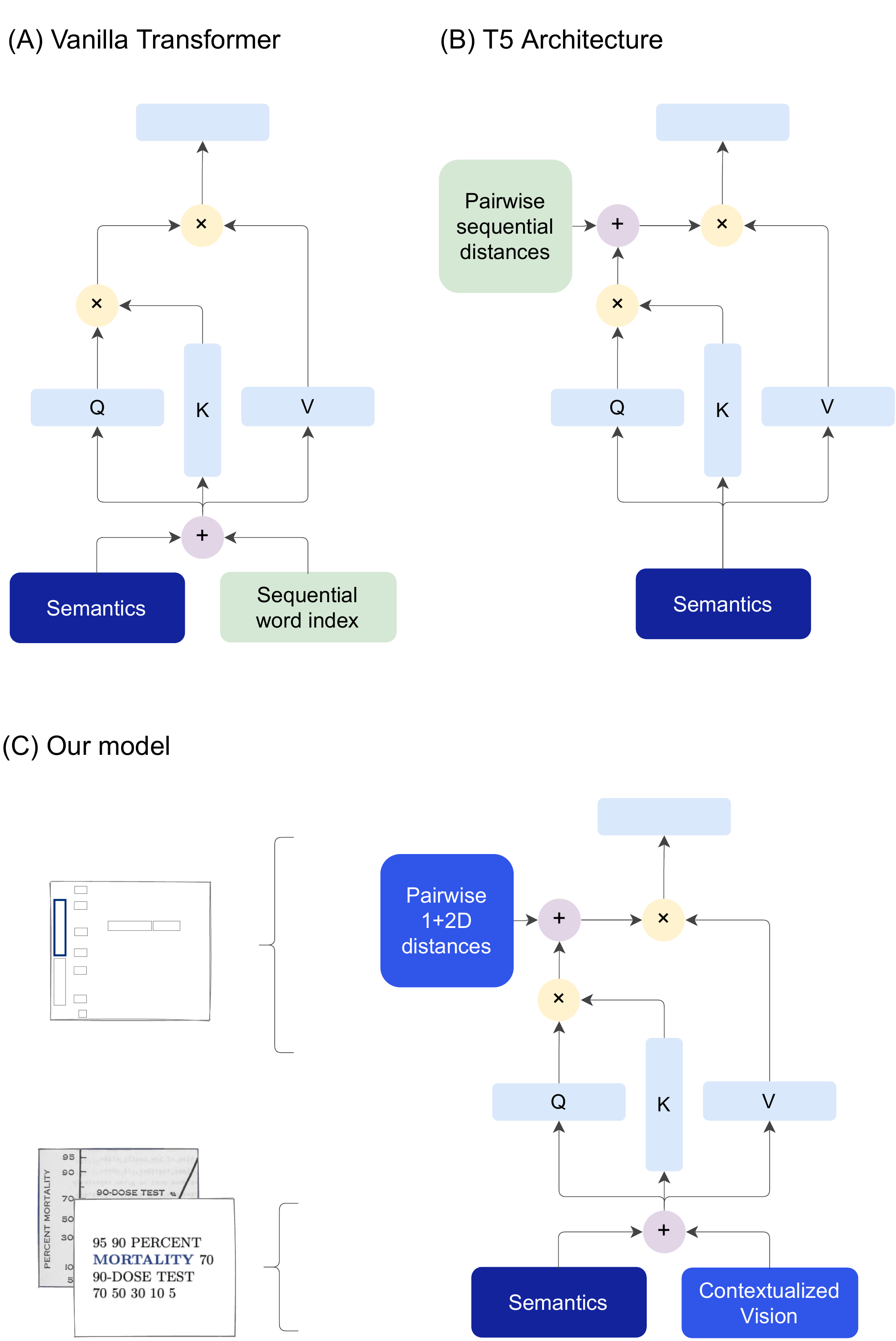}
        \caption{(A)~In the original Transformer, information about the order of tokens is provided explicitly to the model by positional embeddings added to semantic embeddings. (B)~T5 introduces sequential bias, thus separating semantics from sequential distances. (C)~We maintain this clear distinction, extending biases with spatial relationships and providing additional \textit{image semantics} at the input.}
        \label{fig:attention}
    \end{figure}
}{}

\section{Model Architecture}\label{sec:arch}
Our starting point is the architecture of the Transformer, initially proposed for Neural Machine Translation, which has proven to be a solid baseline for all generative tasks involving natural language~\cite{transformer}.

Let us begin from the general view on attention in the first layer of the Transformer.
If $n$ denotes the number of input tokens, resulting in a matrix of embeddings $X$, then self-attention can be seen as:
\begin{equation}
\text{softmax}\left(\frac{Q_X K_X^\top}{\sqrt{n}} + B\right) V_X
\end{equation}
\noindent where $Q_X$, $K_X$ and $V_X$ are projections of $X$ onto query, keys, and value spaces, whereas $B$ stands for an optional attention bias.
There is no $B$ term in the original Transformer, and information about the order of tokens is provided explicitly to the model, that is:
$$X = S + P \hspace{20pt} B = 0_{n\times d}$$
where $S$ and $P$ are respectively the semantic embeddings of tokens and positional embedding resulting from their positions~\cite{transformer}. $0_{n\times d}$ denote a zero matrix.

\ifthenelse{\boolean{true}}{In contrast to the original formulation, we rely on relative attention biases instead of positional embeddings. These are further extended to take into account spatial relationships between tokens (Figure~\ref{fig:attention}).}{}  %, including Grammatical Error Correction~\cite{kiyono-etal-2019-empirical}.

% wychodzimy od gołego transformera i bierzemy najlepsze z innych podejść (embeddingi z T5, lambert?, rezydualna atencja, multimodalność, pretrening na zbiorach o różnej i mieszanej modalności, augmentacje w CV,) + CONTEXTUALIZED REGION EMBEDDING (TM)
% \todo[inline]{relative bias 2d - R}
%\subsection{Base model?}
% We use modified versions of T5 model Raffel~et~al.~\cite{2020t5}

\subsubsection{Spatial Bias.}\label{layout_emb}
%Let $Q$, $K$ and $V$ denote maps $\mathbb{R}^{\ensuremath{\mathsf{dim}}} \rightarrow \mathbb{R}^{\ensuremath{\mathsf{dim}}}$ and $X$ stand for a sequence of input ebeddings $X \in \mathbb{R}^{\ensuremath{\mathsf{seq}} \times \ensuremath{\mathsf{dim}}}$.
%Authors of T5 architecture proposed a strict separation of semantic attention $S$ from one invariant under $S$~\cite{2020t5}. It is achieved by setting $X = S$ and introduction of relational bias, incorporated into self-attention equation through the sequential bias term $B = B^{\text{1D}}$.
%It is a simplified form of positional signal inclusion, where each logit used for computing the attention weights has some learned scalar added, resulting from corresponding token-to-token offsets.
Authors of the T5 architecture disregarded positional embeddings~\cite{2020t5}, by setting $X = S$. They used relative bias by extending self-attention's equation with the sequential bias term $B = B^{\text{1D}}$, a simplified form of positional signal inclusion. Here, each logit used for computing the attention head weights has some learned scalar added, resulting from corresponding token-to-token offsets.

We extended this approach to spatial dimensions. In our approach, biases for relative horizontal and vertical distances between each pair of tokens are calculated and added to the original sequential bias, i.e.: $$B = B^{\text{1D}} + B^{\text{H}} + B^{\text{V}}$$ Such bias falls into one of 32 buckets, which group similarly-distanced token-pairs.
%Such distance is distributed into 32 buckets.
The size of the buckets grows logarithmically so that greater token pair distances are grouped into larger buckets.

\ifthenelse{\boolean{arxiv}}{
    \begin{figure}
        \centering
        \includegraphics[width=0.4\linewidth]{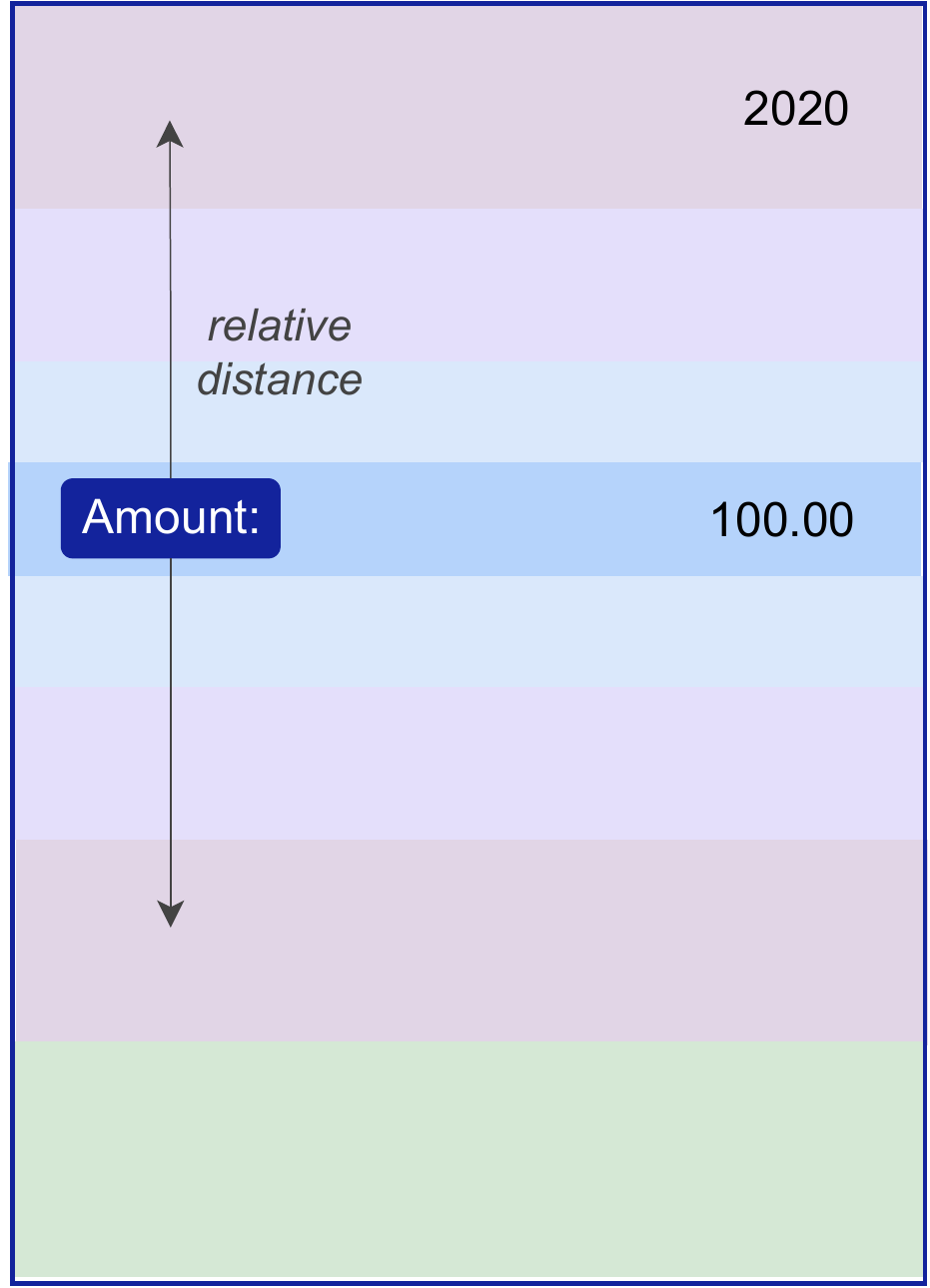}
        \caption{Document excerpt with distinguished vertical buckets for the \textit{Amount} token.}
        \label{fig:vertical_bias}
    \end{figure}
}{
    \begin{figure}%
        \begin{minipage}[b]{0.65\linewidth}%
            \centering%
            \includegraphics[trim=0 0 0 850, clip, width=\linewidth]{images/attention/V3-4.pdf}
            \caption{T5 introduces sequential bias, separating semantics from sequential distances. We maintain this clear distinction, extending biases with spatial relationships and providing additional \textit{image semantics} at the input.}%
            \label{fig:attention}%
        \end{minipage}%
        \hfill
        \begin{minipage}[b]{0.3\linewidth}%
            \centering%
            \includegraphics[width=\linewidth ]{images/spatial-bias/vertical_bias_2.pdf}%
            \caption{Document excerpt with distinguished vertical buckets for the \textit{Amount} token.}%
            \label{fig:vertical_bias}%
        \end{minipage}%
    \end{figure}%
}

%\red{Similar} approach has been \red{simultaneously} applied in layout-aware language modelling by authors of LayoutLMv2~\cite{xu2020layoutlmv2}. \red{Other} authors injected spatial information analogously to how sequential word order is covered in Transformer architecture~\cite{garncarek2020lambert}.

% Propozycja zmiany narracji:
%Other layout-aware language models have spatial information injected analogously to how sequential word order is covered in Transformer architecture~\cite{garncarek2020lambert} or ...~\cite{xu2020layoutlmv2}

\subsubsection{Contextualized Image Embeddings.}\label{sec:image_context}
Contextualized \emph{Word} Embeddings are expected to capture context-dependent semantics and return a sequence of vectors associated with an entire input sequence~\cite{Ethayarajh2019HowCA}. We designed Contextualized \emph{Image} Embeddings with the same objective, i.e., they cover the image region semantics in the context of its entire visual neighborhood.

\begin{figure}
    \centering
    \includegraphics[width=\linewidth]{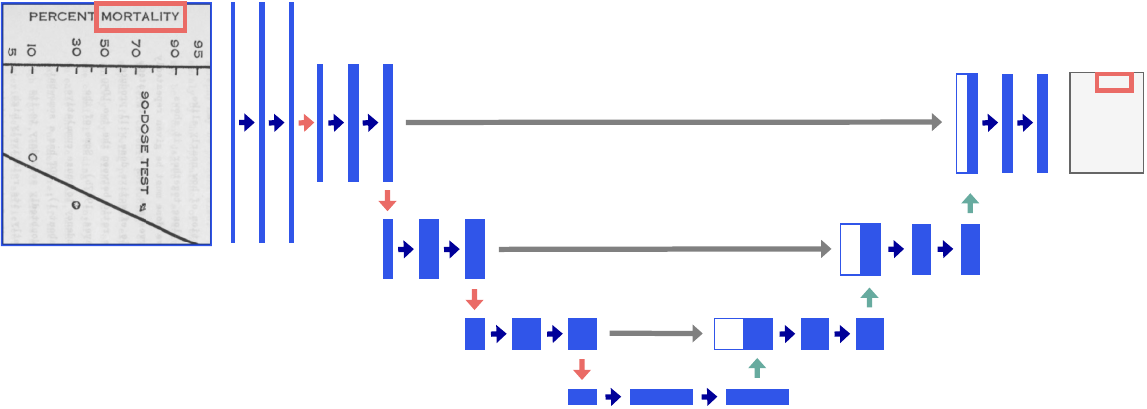}
    \caption{Truncated U-Net network.\quad {\color{applica}{\textbf{$\blacksquare$}}} conv \quad {\color{aorange}{\textbf{$\blacksquare$}}} max-pool \quad {\color{agreen}{\textbf{$\blacksquare$}}} up-conv \quad {\color{agray}{\textbf{$\blacksquare$}}} residual}
    \label{fig:unet}
\end{figure}

%\subsubsection{Visual features.}
To produce image embeddings, we use a convolutional network that consumes the whole page image of size 512×384 and produces a feature map of 64×48×128. We rely on U-Net as a backbone visual encoder network~\cite{RFB15a} since this architecture provides access to not only the information in the near neighborhood of the token, such as font and style but also to more distant regions of the page, which is useful in cases where the text is related to other structures, i.e., is the description of a picture. This multi-scale property emerges from the skip connections within chosen architecture (Figure~\ref{fig:unet}). Then, each token's bounding box is used to extract features from U-Net's feature map with ROI pooling~\cite{dai2016object}. The obtained vector is then fed into a linear layer which projects it to the model embedding dimension.

%\subsubsection{Embeddings.}
In order to inject visual information to the Transformer, a matrix of contextualized image-region embeddings $U$ is added to semantic embeddings, i.e. we define $$X = S + U$$ in line with the convention from Section~\ref{sec:arch} (see Figure~\ref{fig:attention}).

% \todo[inline]{residual attention}

\section{Regularization Techniques}

In the sequence labeling scenario, each document leads to multiple training instances (token classification), whereas in Transformer sequence-to-sequence models, the same document results in one training instance with feature space of higher dimension (decoding from multiple tokens).

Since most of the tokens are irrelevant in the case of Key Information Extraction and contextualized word embeddings are correlated by design, one can suspect our approach to overfit easier than its sequence labeling counterparts. To improve the model's robustness, we introduced a regularization technique for each modality.

\subsubsection{Case Augmentation.}
Subword tokenization~\cite{sennrich-etal-2016-neural,kudo-2018-subword} was proposed to solve the word sparsity problem and keep the vocabulary at a reasonable size. 
Although the algorithm proved its efficiency in many NLP fields, the recent work showed that it performs poorly in the case of an unusual casing of text~\cite{powalski2020unicase}, for instance, when all words are uppercased. The problem occurs more frequently in formated documents (FUNSD, CORD, DocVQA), where the casing is an important visual aspect.
% Subword tokenization~\cite{sennrich-etal-2016-neural,kudo-2018-subword}, commonly used with Transformer architecture, has several identified disadvantages. For example, it is deterministic while it has been shown that non-deterministic segmentation leads to more robust models due to learning the compositionality of words better~\cite{provilkov2020bpedropout}. Moreover, pretrained models tend to underperform when text is written with all capitals since it leads to different segmentation with embeddings of rarely used units~\cite{powalski2020unicase}.
% old(MP_removed): For some of the datasets we extended our train set with different casing variants. Apart from original train examples we use also lower-cased and upper-cased variants for which we transformed both document and target text.
%new(MP_added):
We overcome both problems with a straightforward regularization strategy, i.e., produce augmented copies of data instances by lower-casing or upper-casing both the document and target text simultaneously.

% \todo[inline]{rozszerzyć opis case augmentation - R}

%\todo[inline]{question and document augmentation - Ł?}

\subsubsection{Spatial Bias Augmentation.}
% old(MP_removed):  Following common practices in Computer Vision, where images used for training are randomly transformed, we perform similar transformation on spatial biases. We multiply horizontal and vertical distances between tokens by random factor. Such transformation correspond to stretching or squeezing document page on horizontal and vertical dimension.
%new(MP_added):
Analogously to Computer Vision practices of randomly transforming training images, we augment spatial biases by multiplying the horizontal and vertical distances between tokens by a random factor. Such transformation resembles stretching or squeezing document pages in horizontal and vertical dimensions. Factors used for scaling each dimension were sampled uniformly from range $[0.8, 1.25]$.

\subsubsection{Affine Vision Augmentation.}
To account for visual deformations of real-world documents, we augment images with affine transformation, preserving parallel lines within an image but modifying its position, angle, size, and shear. When we perform such modification to the image, the bounding box of every token is updated accordingly.
The exact hyperparameters were subject to an optimization. We use 0.9 probability of augmenting and report the following boundaries for uniform sampling work best: $[-5, 5]$ degrees for rotation angle, $[-5\%, 5\%]$ for translation amplitude, $[0.9, 1.1]$ for scaling multiplier, $[-5, 5]$ degrees for the shearing angle.

\section{Experiments}

Our model was validated on series of experiments involving Key Information Extraction, Visual Question Answering, classification of rich documents, and Question Answering from layout-rich texts. The following datasets represented the broad spectrum of tasks and were selected for the evaluation process (see Table \ref{tab:datasets-comparison} for additional statistics).

%\subsubsection{Datasets.}
The CORD dataset~\cite{park2019cord} includes images of Indonesian receipts collected from shops and restaurants. The dataset is prepared for the information extraction task and consists of four categories, which fall into thirty subclasses. % The evaluation metric is entity-level F1. Ł: Mamy w tabelce
% \paragraph{SROIE.}
The main goal of the SROIE dataset~\cite{Huang2019ICDAR2019CO} is to extract values for four categories (company, date, address, total) from scanned receipts. % The evaluation metric is F1.
% \paragraph{DocVQA.}
The DocVQA dataset~\cite{mathew2020docvqa} is focused on the visual question answering task. % The evaluation metric is ANLS~\cite{AliFurkanBiten2019}.
% \paragraph{RVL-CDIP.}
The RVL-CDIP dataset~\cite{harley2015icdar} contains gray-scale images and assumes classification into 16 categories such as letter, form, invoice, news article, and scientific publication. % The dataset focuses on a multi-class classification task. %, and the evaluation metric is accuracy.
% \paragraph{WikiTableQuestions.} The key objective of the WikiTableQuestions dataset~\cite{pasupat-liang-2015-compositional} is to answer complex questions on semi-structured HTML tables. There are 22,033 examples in the dataset on 2,108 tables, selected from Wikipedia, with at least 8 rows and 5 columns. In order to answer the questions, multi-step reasoning and various data operations (e.g., comparison, aggregation, arithmetic computation) are required. The evaluation metric is accuracy.
% \paragraph{WikiOps.}
%The WikiOps dataset~\cite{Cho2018AdversarialTA} consists of tables extracted from Wikipedia and natural language questions corresponding to them. % SQL queries have been prepared for each table.
%Each has an operand information assigned. %The main purpose of this dataset is to answer the given questions. %, and the evaluation metric is accuracy.
For DocVQA, we relied on Amazon Textract OCR; for RVL-CDIP, we used Microsoft Azure OCR, for SROIE and CORD, we depended on the original OCR.
\begin{table}[ht!]
    \caption{Comparison of datasets considered for supervised pretraining and evaluation process. Statistics given in thousands of documents or questions.\label{tab:datasets-comparison}}
    \setlength{\tabcolsep}{6pt}
    \centering
    \begin{tabular}{llcrr}
        \toprule
        Dataset &
        Data type &
        Image &
        Docs (k) &
        Questions (k) \\
        \midrule
        CORD~\cite{park2019cord} & receipts & $+$ & 1.0 & --- \\
        
        SROIE~\cite{Huang2019ICDAR2019CO} & receipts & $+$ & 0.9 & --- \\
        
        DocVQA~\cite{mathew2020docvqa} & industry documents & $+$ & 12.7 & 50.0 \\
        
        RVL-CDIP~\cite{harley2015icdar} & industry documents & $+$ & 400.0 & --- \\
        \midrule
        DROP~\cite{Dua2019DROPAR} & \rdelim\}{5}{30pt}[ Wikipedia pages] & $-$ & 6.7 & 96.5  \\

        QuAC~\cite{DBLP:journals/corr/abs-1808-07036} &  &  $-$ &  13.6 & 98.4 \\

        SQuAD 1.1~\cite{DBLP:journals/corr/RajpurkarZLL16} &  &  $-$ & 23.2 & 107.8 \\

        TyDi QA~\cite{tydiqa} &  &  $-$ & 204.3 & 204.3 \\

        Natural Questions~\cite{47761} &  & $-$ & 91.2 & 111.2\vspace{1mm}  \\

        WikiOps~\cite{Cho2018AdversarialTA} & Wikipedia tables & $-$ & 24.2 & 80.7 \\

        CoQA~\cite{DBLP:journals/corr/abs-1808-07042} & various sources &  $-$ & 8.4 & 127.0 \\

        RACE~\cite{lai2017large} & English exams &  $-$ & 27.9 & 97.7 \\

        QASC~\cite{Khot2020QASCAD} & school-level science &  $-$ & --- & 10.0 \\

        FUNSD~\cite{jaume2019} & RVL-CDIP forms &  $+$ & 0.1 & --- \\

        Infographics VQA & infographics &  $+$ & 4.4 & 23.9 \\

        TextCaps~\cite{sidorov2019textcaps} & Open Images &  $+$ & 28.4 & --- \\

        DVQA~\cite{kafle2018dvqa} & synthetic bar charts &  $+$ & 300.0 & 3487.2 \\

        FigureQA~\cite{Kahou2018FigureQAAA} & synthetic, scientific &  $+$ & 140.0 & 1800.0 \\

        TextVQA~\cite{singh2019towards} & Open Images &  $+$ & 28.4 & 45.3 \\
        \bottomrule \\
    \end{tabular}
\end{table}

\subsection{Training Procedure}\label{sec:training_procedure}
The training procedure consists of three steps. First, the model is initialized with vanilla T5 model weights and is pretrained on numerous documents in an unsupervised manner. It is followed by training on a set of selected supervised tasks. Finally, the model is finetuned solely on the dataset of interest. We trained two size variants of TILT models, starting from T5-Base and T5-Large models. Our models grew to 230M and 780M parameters due to the addition of Visual Encoder weights.

%\todo[inline]{Datasets (wr, syntetic 2d, salient spans) - D + T + G}
\subsubsection{Unsupervised Pretraining.} We constructed a corpus of documents with rich structure, based on RVL-CDIP ($275$k docs), UCSF Industry Documents Library ($480$k),\footnote{\url{http://www.industrydocuments.ucsf.edu/}} and PDF files from Common Crawl ($350$k). The latter were filtered according to the score obtained from a simple SVM business document classifier.

Then, a T5-like masked language model pretraining objective is used, but in a salient span masking scheme, i.e., named entities are preferred rather than random tokens~\cite{2020t5,guu2020realm}.
Additionally, regions in the image corresponding to the randomly selected text tokens are masked with the probability of $80\%$. Models are trained for $100,000$ steps with batch size of $64$, AdamW optimizer and linear scheduler with an initial learning rate of $2e-4$.

\subsubsection{Supervised Training.}
%Pre-training on a set of similar tasks is a common approach in transfer-learning that allows obtain\red{ing} higher scores on a particular task.
%Therefore, we chose sixteen datasets that the final task could benefit from.
%Table \ref{tab:datasets-comparison} presents a basic comparison between the datasets.
%The datasets differ between each other in terms of size, domain, and information about document structure.
%In case of pure text dataset, we imitated the structure by converting text to a HTML or PDF document.

\begin{figure}[htp]
\centering
\includegraphics[trim=0 40 0 
130,clip,width=0.9\textwidth]{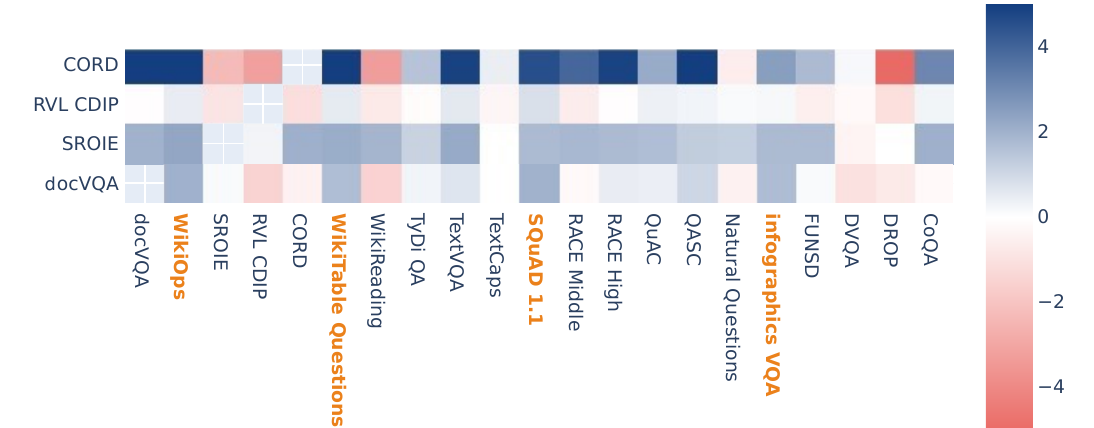}\hfill
\caption{Scores on CORD, DocVQA, SROIE and RVL-CDIP compared to the baseline without supervised pretraining. % on four datasets: CORD, docVQA, RVL-CDIP, WikiOps (columns).
The numbers represent the differences in the metrics, %: F1 score (CORD and RVL-CDIP), Accuracy (WikiOps), and ANLS (docVQA)
orange text denote datasets chosen for the final supervised pretraining run.}
\label{fig:pretraining}
\end{figure}

To obtain a general-purpose model which can reason about documents with rich layout features, we constructed a dataset relying on a large group of tasks, representing diverse types of information conveyed by a document (see Table \ref{tab:datasets-comparison} for datasets comparison). Datasets, which initially had been plain-text, had their layout produced, assuming some arbitrary font size and document dimensions. Some datasets, such as \textit{WikiTable Questions}, come with original HTML code -- for the others, we render text alike. Finally, an image and computed bounding boxes of all words are used.

At this stage, the model is trained on each dataset for 10,000 steps or 5 epochs, depending on the dataset size: the goal of the latter condition was to avoid a quick overfitting.

%\todo{jak tworzymy input dla modelu (prefix)}

% The selection process \red{was} based on the following procedure.
% We trained a model on each dataset separately for 10,000 updates or ten epochs and then fine-tuned on the particular evaluation dataset.
% In addition, we performed the procedure on the concatenation of all the datasets.
% As a baseline, we trained a model on the desired dataset for the same number of updates.
% For the final training, we chose those datasets that  performed significantly better than the baseline.
% The final pretraining set consists of the chosen datasets and the dataset for the desired task.
We estimated each dataset's value concerning a downstream task, assuming a fixed number of pretraining steps followed by finetuning. The results of this investigation are demonstrated in Figure~\ref{fig:pretraining}, where the group of WikiTable, WikiOps, SQuAD, and infographicsVQA performed robustly, convincing us to rely on them as a solid foundation for further experiments.

Model pretrained in unsupervised, and then supervised manner, is at the end finetuned for two epochs on a downstream task with AdamW optimizer and hyperparameters presented in Table~\ref{tab:finetune-params}.

\begin{table}
    \caption{Parameters used during the finetuning on a downstream task. Batch size, learning rate and scheduler were subject of hyperparameter search with considered values of respectively $\{8, 16, ..., 2048\}$, $\{\mathrm{5e-5, 2e-5, 1e-5, 5e-4, ..., 1e-3}\}$, $\{\mathrm{constant, linear}\}$. We have noticed that the classification task of RVL-CDIP requires a significantly larger bath size. The model with the highest validation score within the specified steps number limit was used. \label{tab:finetune-params}}
    \setlength{\tabcolsep}{5pt}
    \centering
    \begin{tabular}{*7c}
        \toprule
        Dataset & Batch size & Steps & Learning rate & Scheduler \\
        \midrule
        SROIE & 8 & 6,200 & 1e-4 & constant  \\
        % WikiOps & 64 & 4,200 & 1e-4 & constant  \\
        DocVQA & 64 & 100,000 & 2e-4 & linear  \\
        CORD & 8 & 36,000 & 2e-4 & linear  \\
        RVL-CDIP & 1,024 & 12,000 & 1e-3 & linear  \\
        \bottomrule  \\
    \end{tabular} 
\end{table}

\subsection{Results}

The TILT model achieved state-of-the-art results on three out of four considered tasks (Table~\ref{tab:results}). We have confirmed that unsupervised layout- and vision-aware pretraining leads to good performance on downstream tasks that require comprehension of tables and other structures within the documents. Additionally, we successfully leveraged supervised training from both plain-text datasets and these involving layout information.

\begin{table}[t!]
    \caption{Results of selected methods in relation to our base and large models. Bold indicates the best score in each category. All results on the test set, using the metrics proposed by dataset's authors. The number of parameters given for completeness thought encoder-decoder and LMs cannot be directly compared under this criterion.  \label{tab:results}}
    \setlength{\tabcolsep}{5.5pt}
    \centering
    \begin{tabular}{lccccl}
        \toprule
        \multirow{2}{*}{Model} &
        CORD &
        SROIE &
        DocVQA &
        %WikiOps &
        RVL-CDIP & Size variant \\
        & F1 & F1 & ANLS 
        % Accuracy
        & Accuracy & (Parameters) \\
        \midrule
         \multirow{2}{*}{LayoutLM~\cite{xu2020layoutlm}} & $94.72$ & $94.38$ & $69.79$ & $94.42$ & Base\hspace{2.4mm}(113-160M) \\
         & $94.93$ & $95.24$ & $72.59$ & $94.43$ & Large (343M) \\
         \multirow{2}{*}{LayoutLMv2~\cite{xu2020layoutlmv2}} & $94.95$ & $96.25$ & $78.08$ & $95.25$ & Base\hspace{2.4mm}(200M) \\
         & $96.01$ & $97.81$ & $86.72$ & $\mathbf{95.64}$ & Large (426M) \\
        LAMBERT~\cite{garncarek2020lambert} & $96.06$ & $\mathbf{98.17}$ & --- & --- & Base\hspace{2.4mm}(125M) \\
        % BROS~\cite{hong2021bros} & & $95.36$ & $95.48$ & --- & --- & Unknown \\
        % NeOp~\cite{Cho2018AdversarialTA} & --- & --- & --- & $59.50$ & ---
        \vspace{0mm} \\
        \multirow{2}{*}{TILT (our)} & $95.11$ & $97.65$ & $83.92$ & %$69.16$ &
        $95.25$ & Base\hspace{2.4mm}(230M) \\
         & $\mathbf{96.33}$ & $\mathbf{98.10}$ & $\mathbf{87.05}$ %& $\mathbf{73.80}$
        & $95.52$ & Large (780M) \\
        \bottomrule \\
    \end{tabular}
\end{table}

\subsubsection{DocVQA.}
We improved SOTA results on this dataset by $0.33$ points. 
Moreover, detailed results show that model gained the most in table-like categories, i.e., forms ($89.5$ \textrightarrow~$94.6$) and tables ($87.7$ \textrightarrow~$89.8$), which proved its ability to understand the spatial structure of the document.
Besides, we see a vast improvement in the yes/no category ($55.2$ \textrightarrow~$69.0$).\footnote{Per-category test set scores are available after submission on the competition web page: \ifthenelse{\boolean{true}}{\url{https://rrc.cvc.uab.es/?ch=17\&com=evaluation\&task=1}}{\textit{(Link removed due to anonimity concerns)}}.}
In such a case, our architecture generates simply \textit{yes} or \textit{no} answer, while sequence labeling based models require additional components such as an extra classification head.
We noticed that model achieved lower results in the image/photo category, which can be explained by the low presence of image-rich documents in our datasets.

\subsubsection{RVL-CDIP.}
Part of the documents to classify does not contain any readable text.
Because of this shortcoming, we decided to guarantee there are at least 16 image tokens that would carry general image information.
Precisely, we act as there were tokens with bounding boxes covering 16 adjacent parts of the document. These have representations from U-Net, exactly as they were regular text tokens. Our model places second, $0.12$ below the best model, achieving the similar accuracy of $95.52$. % We consider it more than satisfactory. % since the best image-based classification method so far is several points worse ($92.31$), and our score is close to the dataset's upper bound~\cite{Ferrando2020ImprovingAA}. Moreover, we presented the best-performing model that does not rely on a specialized document classification head.  %\todo{Scores, beauty}

% For RVL which is a classification task, overall understanding of the image might be beneficial.
% The beauty of our approach allowed us to inject whole image information without changing the architecture of our model.
% We decided to add 16 extra image tokens\footnote{For each image token we used \texttt{<extra\_token\_id\_99>} sentinel token from T5.} before the input prefix that would carry general image information.
% We split image into 16 parts and added ROI pooled representations from U-Net to the image tokens, exactly like it was done for normal text tokens. Same was done with vertical and horizontal biases which were calculated by relying on coordinates of those 16 image parts.

\subsubsection{CORD.}
Since the complete inventory of entities is not present in all examples, we force the model to generate a \textit{None} output for missing entities.
Our model achieved SOTA results on this challenge and improved the previous best score by $0.3$ points.
Moreover, after the manual review of the model errors, we noticed that model's score could be higher since the model output and the reference differ insignificantly e.g. "2.00 ITEMS" and "2.00".

\subsubsection{SROIE.}
We excluded OCR mismatches and fixed total entity annotations discrepancies following the same evaluation procedure as Garncarek et~al.~\cite{garncarek2020lambert}.\footnote{Corrections can be obtained by comparing their two public submissions.} We achieved results indistinguishable from the SOTA ($98.10$ vs. $98.17$). Significantly better results are impossible due to OCR mismatches in the test-set.

\bigskip\noindent Though we report the number of parameters near the name of the model size variant, note it is impossible to compare the TILT encoder-decoder model to language models such as LayoutLMs and LAMBERT under this criterion. In particular, it does not reflect computational cost, which may be similar for encoder-decoders twice as big as some language model \cite[Section 3.2.2]{2020t5}. Nevertheless, it is worth noting that our Base model outperformed models with comparable parameter count.

%\todo[inline]{Methodology}

\section{Ablation study}
In the following section, we analyze the design choices in our architecture, considering the base model pretrained in an unsupervised manner and the same hyperparameters for each run.
The DocVQA was used as the most representative and challenging for Document Intelligence since its leaderboard reveals a large gap to human performance.
We report average results over two runs of each model varying only in the initial random seed to account for the impact of different initialization and data order~\cite{dodge2020finetuning}. % We report the ANLS metric on the validation set.

% Version 1             WIP
% \begin{table}[b]
%     \setlength{\tabcolsep}{7pt}
%     \centering
%     \begin{tabular}{cccccc}
%         \toprule
%         \multirow{2}{*}{2D bias} &
%         \multirow{2}{*}{Visual embeddings} &
%         \multicolumn{3}{c}{Augmentation} &
%         \multirow{2}{*}{Score} \\
%         & & Case & Spatial & Vision & \\
%         \midrule
%         $+$ & $+$ & $+$ & $+$ & $+$ & $0.0 \pm 0.0$ \\
%         $-$ & $+$ & $+$ & $+$ & $+$ & $0.0 \pm 0.0$ \\
%         $+$ & $-$ & $+$ & $+$ & $+$ & $0.0 \pm 0.0$ \\
%         $+$ & $+$ & $-$ & $+$ & $+$ & $0.0 \pm 0.0$ \\
%         $+$ & $+$ & $+$ & $-$ & $+$ & $0.0 \pm 0.0$ \\
%         $+$ & $+$ & $+$ & $+$ & $-$ & $0.0 \pm 0.0$ \\
%         \bottomrule  \\
%     \end{tabular}
%     \caption{Ablation study. \label{tab:ablation}}
% \end{table}

% Version 2             WIP
\begin{table}[t]
    \caption{Results of ablation study. The minus sign indicates removal of the mentioned part from the base model. \label{tab:ablation}}
    \setlength{\tabcolsep}{7pt}
    \centering
    \begin{tabular}{lcc}
        \toprule
        Model &
        Score &
        Relative change\\

        \midrule
        TILT-Base & $82.9 \pm 0.3$ & ---  \\
        \quad -- Spatial Bias & $81.1 \pm 0.2$ & $-1.8$\\
        \quad -- Visual Embeddings& $81.2 \pm 0.3$ & $-1.7$\\
        \quad -- Case Augmentation & $82.2 \pm 0.3$ & $-0.7$\\
        \quad -- Spatial Augmentation & $82.6 \pm 0.4$ & $-0.3$\\
        \quad -- Vision Augmentation & $82.8 \pm 0.2$ & $-0.1$\\
        \quad -- Supervised Pretraining & $81.2 \pm 0.1$ & $-1.7$\\
        \bottomrule  \\
    \end{tabular}
\end{table}

\subsubsection{Significance of Modalities.} We start with the removal of the 2D layout positional bias. Table~\ref{tab:ablation} demonstrates that information that allows models to recognize spatial relations between tokens is a crucial part of our architecture. It is consistent with the previous works on layout understanding~\cite{xu2020layoutlmv2,garncarek2020lambert}. 
Removal of the UNet-based convolutional feature extractor results in a less significant ANLS decrease than the 2D bias. This permits the conclusion that contextualized image embeddings are beneficial to the encoder-decoder.

\subsubsection{Justifying Regularization.} Aside from removing modalities from the network, we can also exclude regularization techniques.
To our surprise, the results suggest that the removal of case augmentation decreases performance most severely. Our baseline is almost one point better than the equivalent non-augmented model.
Simultaneously, model performance tends to be reasonably insensitive to the bounding boxes' and image alterations.
%\todo{write about image augmentation}
It was confirmed that other modalities are essential for the model's success on real-world data, whereas regularization techniques we propose slightly improve the results, as they prevent overfitting.

\subsubsection{Impact of Pretraining.} As we exploited supervised pretraining similarly to previous authors, it is worth considering its impact on the overall score. In our ablation study, the model pretreated in an unsupervised manner achieved significantly lower scores. The impact of this change is comparable to the removal of spatial bias or visual embeddings. Since authors of the T5 argued that pretraining on a mixture of unsupervised and supervised tasks perform equally good with higher parameter count, this gap may vanish with larger variants of TILT we did not consider in the present paper \cite{2020t5}.

% many additional perspectives like having visual features, 1-, and 2-dimensional layout information and textual information are essential to making IE systems work well on real data
%     \item that model regularization techniques improve results, preventing overfitting in the training of large models

% all results - the peak anls result on docvqa
% baseline (unet+2d+caug+baug) - 82.84 82.76
% wo unet 81.22
% wo 2d 81.09
% wo caug 82.15
% wo baug 82.59 82.60

% extra
% add imgaug 83.05 82.75

% \section{Discussion}

% We have confirmed that unsupervised layout- and vision-aware pretraining leads to good performance on downstream tasks that require comprehension of tables and other structures within the documents. Additionally, we successfully leveraged supervised training from both plain-text datasets and these involving layout information.
% \begin{enumerate}
%     \item that unsupervised layout-pretraining on various documents effectively handles the challenge of recognition different structures in the document
%     \item that a combination of question answering and layout-focused datasets can be used to train state-of-the-art Key Information Extraction models consistently
%     \item that many additional perspectives like having visual features, 1-, and 2-dimensional layout information and textual information are essential to making IE systems work well on real data
%     \item that model regularization techniques improve results, preventing overfitting in the training of large models
% \end{enumerate}

\section{Summary}

In the present paper, we introduced a novel encoder-decoder framework for layout-aware models. Compared to the sequence labeling approach, the proposed method achieves better results while operating in an end-to-end manner. It can handle various tasks such as Key Information Extraction, Question Answering or Document Classification, while the need for complicated preprocessing and postprocessing steps is eliminated.

Although encoder-decoder models are commonly applied to generative tasks, both DocVQA, SROIE, and CORD we considered are extractive. We argue that better results were achieved partially due to the independence from the detected word order and resistance to OCR errors that the proposed architecture possesses. 
Consequently, we were able to achieve state-of-the-art results on two datasets (DocVQA, CORD) and performed on par with the previous best scores on SROIE and RVL-CDIP, albeit having a much simpler workflow.

Spatial and image enrichment of the Transformer model allowed the TILT to combine information from text, layout, and image modalities. We showed that the proposed regularization methods significantly improve the results.

\ifthenelse{\boolean{true}}{\subsubsection{Acknowledgments.} The authors would like to thank Filip Graliński, Tomasz Stanisławek, and Łukasz Garncarek for fruitful discussions regarding the paper and our managing directors at Applica.ai. Moreover, Dawid Jurkiewicz pays due thanks to his son for minding the deadline and generously coming into the world a day after.\\\\The Smart Growth Operational Programme supported this research under project no. POIR.01.01.01-00-0877/19-00 (\textit{A universal platform for robotic automation of processes requiring text comprehension, with a unique level of implementation and service automation}).}{}

% We have shown that varied tasks involving processing of visually-rich documents can be unified and solved using the encoder-decoder architecture of our design.

% This new encoder-decoder framework for layout-aware models comes with advantages and disadvantages relative to the previous sequence-labeling works. The primary disadvantage of the technique is the autoregressive nature of answer generation. Although the decoder's depth can be lowered to alleviate this, much of the answers are below XXX tokens in length.
% However, with the enhancement of training in an encoder-decoder fashion, no assumptions about problem structure are made, suggesting that our approach will likely do well on other challenging Document Visual Question Answering tasks. Our method can be employed without tagging, normalizing, and aggregating, as it is feasible to train end-to-end -- something that sequence-labeling baseline models cannot do directly. Even more importantly, the encoder-decoder setup outperforms the baselines on the exact match(?) subgroup - to which both the models can be applied.
% As a consequence of our method's advances, a new class of information extraction problems that can be tackled without artificial assumptions was opened.
% We demonstrated the soundness of this approach on the public benchmarks of DocVQA, CORD, SROIE, WikiOps, RVL-CDIP, where our system's extraction quality exceeds all published results.


\bibliographystyle{splncs04}
\bibliography{bibliography}

% \clearpage
% \input{appendix}
\end{document}